\documentclass[11pt,a4paper]{article}
\usepackage[hyperref]{acl2018}
\usepackage{times}
\usepackage{latexsym}
\usepackage{footnote}
\usepackage{tablefootnote}
\usepackage{enumitem}

\usepackage{bm}
\usepackage{multirow}
\usepackage{amsfonts}
\usepackage[boxed,linesnumbered,lined]{algorithm2e}
\usepackage{algpseudocode}
\usepackage{graphicx}
\usepackage{hyphenat}
\usepackage{verbatim}

\usepackage{CJKutf8}
\usepackage{graphicx}
\usepackage{nonfloat}
\usepackage{fmtcount}
\usepackage{threeparttable}
\usepackage{caption}
\usepackage{amsmath}

\usepackage{url}

\aclfinalcopy %

\newcommand{\contest}{CAIL2018 }
\newcommand{\task}{LJP }

\title{Overview of CAIL2018: Legal Judgment Prediction Competition}

\author{Haoxi Zhong$^{1\ast}$ Chaojun Xiao$^{1}$\thanks{$ $ $ $ $ $ indicates equal contribution.} $ $ Zhipeng Guo$^{1}$ Cunchao Tu$^{1}$ Zhiyuan Liu$^{1}$ \\ \textbf{Maosong Sun$^{1}$ Yansong Feng$^{2}$ Xianpei Han$^{3}$ Zhen Hu$^{4}$ Heng Wang$^4$ Jianfeng Xu$^{5}$}\\
\normalsize $^{1}$Department of Computer Science and Technology, Tsinghua University, China \\
\normalsize $^{2}$Institute of Computer Science and Technology, Peking University, China\\
\normalsize $^{3}$Institute of Software, Chinese Academy of Sciences, China\\
\normalsize $^{4}$China Justice Big Data Institute\\
\normalsize $^{5}$Supreme People’s Court, China\\
}
\date{}

\begin{document}
\begin{CJK*}{UTF8}{gbsn}

\maketitle
\begin{abstract}
In this paper, we give an overview of the \textbf{L}egal \textbf{J}udgment \textbf{P}rediction  (LJP) competition at \textbf{C}hinese \textbf{AI} and \textbf{L}aw challenge  (CAIL2018). This year's competition focuses on LJP which aims to predict the judgment results according to the given facts. Specifically, in \contest, we proposed three subtasks of \task for the contestants, i.e., predicting relevant law articles, charges and prison terms given the fact descriptions. CAIL2018 has attracted several hundreds participants  ($601$ teams, $1,144$ contestants from $269$ organizations). In this paper, we provide a detailed overview of the task definition, related works, outstanding methods and competition results in CAIL2018.

\end{abstract}

\section{Introduction} \label{section:intro}
Legal Judgment Prediction is a traditional task in the combination of artificial intelligence and laws. It aims to train a machine judge to predict the judgment results (e.g., relevant law articles, charges, prison terms and so on) automatically according to the facts. A well-performed LJP system can not only benefit those who are not familiar with laws but also provide a reference to professionals, e.g., lawyers and judges.

In order to promote the development of legal intelligence, this year's AI and Law challenge, \contest, focuses on how artificial intelligence can help the LJP system. Firstly, we published a large-scale criminal dataset constructed from Chinese law documents~\cite{xiao2018cail2018}. Based on this dataset, we propose three subtasks of LJP for contestants, including predicting relevant law articles, charges and prison terms given the fact descriptions from law documents. 

The goal of \contest is to explore how NLP techniques and legal knowledge benefit the performance of LJP.  For the three subtasks in LJP, there are several major challenges for contestants as follows:

\begin{itemize}
    \item The distributions of various law articles, charges, and prison terms are quite imbalanced. According to the statistics，the top-$10$ charges covers over $79.0\%$ cases while the bottom-$10$ charges only cover about $0.12\%$ cases. The imbalanced distribution makes it difficult to predict low-frequency categories.
    \item Predicting the prison terms via the fact descriptions is more challenging than other subtasks. In real-world scenarios, when deciding the prison terms of a case, the judge will be affected by plenty of factors, e.g., ages of defendants, amount of money involved in the case and so on. It's challenging for a machine to define and extract sufficient features from fact description.
   \item There are usually complex logic dependencies between subtasks. For example, the charges of the criminals should refer to the relevant articles as in Chinese Criminal Law, and the decision of prison terms should accord with the stipulations in law articles. So it is crucial for the contestants to understand the rules contained in law articles and discover the logic dependencies among subtasks.
    \item There exists many confusing categories pairs in these subtasks, such as the charges of robbery and theft. In Chinese Criminal Law, there are only a few differences between the definitions of many charge pairs, which make it difficult to distinguish these confusing charges.
\end{itemize}

In this year's competition, there are $202$ teams who have submitted their models to the contests, and the best-performed models reach $90.62, 87.91, 78.22$ in the three subtasks. Comparing with the performance at the early stage of this competition, all the subtasks achieve significant improvements.

In the following parts, we will give a detailed introduction to \contest including the task definition and evaluation metrics. In addition, we will introduce the best-performed models submitted by contestants and discuss the reminding challenges.



\section{Related Work}
LJP is a hot topic in the field of legal intelligence and has been studied for several years. In early years, the studies of LJP usually concentrate on how to utilize mathematical and statistical methods to build LJP systems in some specific scenarios~\cite{kort1957predicting,ulmer1963quantitative,nagel1963applying,keown1980mathematical,segal1984predicting,lauderdale2012supreme}.

With the development of machine learning techniques, more works propose to employ existing machine learning models to improve the performance on LJP. In these works, they usually formalize LJP as a text classification problem and focus on extracting efficient shallow features from the given facts and additional resources~\cite{liu2006exploring,lin2012,aletras2016predicting,Sulea2017Predicting}. These works integrate machine learning methods into LJP tasks and achieve a promising performance of LJP. However, these conventional methods can only extract well-defined shallow textual features from the fact descriptions.

In recent years, with the successful usage of deep learning techniques on NLP tasks~\cite{kim2014convolutional, Baharudin2010A,tang2015document}, researchers propose to employ neural models to solve LJP tasks. For example, \citet{luo2017learning} adopt attention mechanism between facts and relevant law articles for charge prediction. \citet{niubi2018coling} introduce several charge attributes to predict few-shot and confusing charges. \citet{jiang2018interpretable} employ deep reinforcement learning to extract rationales for interpretable charge prediction. \citet{zhong2018topjudge} model the dependencies among the different subtasks in LJP as a Directed Acyclic Graph (DAG), and propose a topological learning model to solve these tasks simultaneously. \citet{Ye2018Interpretable} integrate Seq2Seq model and predicted charges to generate the court view with fact descriptions. 

\section{Task Definition and Evaluation Metrics}

In this section, we give the detailed dataset construction, task definition, and evaluation metrics of this competition. All the details can also be achieved from \url{https://github.com/thunlp/CAIL}.

\subsection{Dataset Construction}
We construct the CAIL2018 dataset from $5,730,302$ criminal documents collected from China Judgment Online\footnote{\url{http://wenshu.court.gov.cn/}}. As all the law documents are written in a standard format, it is easy to extract the fact description and the judgment results from these documents. During the preprocessing period, we filter out some case documents with low-frequency categories or multiple defendants. Finally, there are $183$ different criminal law articles and $202$ different charges in this dataset.

We randomly selected $1,710,856$ documents as the training set. There are two stages in the contest. In the first stage, we selected $217,016$ documents for testing. After all participants confirmed their final models, we collected $35,922$ emerging documents for testing in the second stage.


\subsection{Task Definition}
LJP takes the fact description of a specific case as the input and predicts the judgment results as the output. The judgment results consist of three parts as follows:

\begin{itemize}
    \item \emph{Law articles}. The contestants should give a list of relevant articles as there might be multiple law articles relevant to one case. 
    \item \emph{Charges}. The contestants should give a list of charges that the defendant in the case is convicted of.
    \item \emph{Prison terms}. The contestants should give the prison term that the defendant in the case is sentenced to. The prison terms should be an integer which stands for how much months the prison terms should be.
\end{itemize}

We denote the prediction of law articles, charges, and prison terms as task $1$, $2$, and $3$ respectively.


\subsection{Evaluation Metrics}
For task $1$ and task $2$, we take them as text classification problems. For a specific task, suppose there are $N$ categories and $M$ documents in total. We denote the ground truth category as $y$ and the predicted label as $\bar{y}$. If the $j$-th documents are annotated with the $i$-th category, then $y_{ij}$ should be $1$ and $0$ otherwise. Then we can get the following metrics for all classes:

\begin{equation}
\begin{aligned}
    \text{TP}_i &= \sum_{j=1}^M [y_{ij}=1,\bar{y}_{ij}=1],\\
    \text{FP}_i &= \sum_{j=1}^M [y_{ij}=0,\bar{y}_{ij}=1],\\
    \text{FN}_i &= \sum_{j=1}^M [y_{ij}=1,\bar{y}_{ij}=0],\\
    \text{TN}_i &= \sum_{j=1}^M [y_{ij}=0,\bar{y}_{ij}=0].\\
\end{aligned}
\end{equation}


These four metrics represent the true positive, false positive, false negative and true negative value for the $i$-th category. Then we can calculate the precision, recall and $F$ value for the $i$-th category as follows:

\begin{equation}
\begin{aligned}
    P_{i} &= \frac{\text{TP}_{i}}{\text{TP}_{i} + \text{FP}_{i}}, \\
    R_{i} &= \frac{\text{TP}_{i}}{\text{TP}_{i} + \text{FN}_{i}},\\
    F_{i} &= \frac{2 \times P_{i} \times R_{i}}{P_{i} +R_{i}}.\\
\end{aligned}
\end{equation}
Here, $p$ and $r$ represent precision and recall respectively. With these evaluation results for all categories, we can calculate the macro-level $F$ value as follows:
\begin{equation}
    F_{\text{macro}} = \frac{\sum_{i=1}^{N}F_{i}}{N}
\end{equation}

Besides, we also evaluate the performance in micro-level. For micro-level evaluation, we first calculate:
\begin{equation}
\begin{aligned}
    \text{TP}_{\text{micro}} &= \sum_{i=1}^{N}\text{TP}_{i},\\
    \text{FP}_{\text{micro}} &= \sum_{i=1}^{N}\text{FP}_{i},\\
    \text{FN}_{\text{micro}} &= \sum_{i=1}^{N}\text{FN}_{i}.\\
\end{aligned}
\end{equation}

Similarly, we can calculate the precision, recall, and $F$ values in the micro-level as follows:

\begin{equation}
\begin{aligned}
    P_{\text{micro}} &= \frac{\text{TP}_{\text{micro}}}{\text{TP}_{\text{micro}} + \text{FP}_{\text{micro}}},\\
    R_{\text{micro}} &= \frac{\text{TP}_{\text{micro}}}{\text{TP}_{\text{micro}} + \text{FN}_{\text{micro}}},\\
    F_{\text{micro}} &= \frac{2 \times P_{\text{micro}} \times R_{\text{micro}}}{P_{\text{micro}} + R_{\text{micro}}}.\\
\end{aligned}
\end{equation}

Finally, we calculate overall score $S$ as

\begin{equation}
    S = 100 \times \frac{F_{\text{micro}} + F_{\text{macro}}}{2}
\end{equation}

For task $3$, we employ the difference of the predicted prison terms and the ground-truth ones as the evaluation metric. Assume that the ground-truth prison term of the $i$-th case is $t_i$ and the predicted result is $\bar{t}_i$. Then, we define the difference $d_i$ as

\begin{equation}
    d_i = \left| \log (t_i + 1) - \log (\bar{t}_i + 1)\right|.
\end{equation}

After that, we define the score function $f(v)$ as :

\begin{equation}
    f (v)=\left\{
    \begin{matrix}
        1.0, & \text{if }v\leq 0.2,\\
        0.8, & \text{if }0.2<v\leq 0.4,\\
        0.6, & \text{if }0.4<v\leq 0.6,\\
        0.4, & \text{if }0.6<v\leq 0.8,\\
        0.2, & \text{if }0.8<v\leq 1,\\
        0.0, & \text{if }1<v.\\
    \end{matrix}
    \right.
\end{equation}

Then the final score of task $3$ should be:

\begin{equation}
    \text{S}=\sum_{i=1}^M\frac{f (d_i)}{M}
\end{equation}


\section{Approach Overview}

\begin{table*}[htb]
\centering
\begin{tabular}{c|cc|cc|c}
\textbf{Tasks}  & \multicolumn{2}{c|}{Law Articles} & \multicolumn{2}{c|}{Charges} & Prison Terms \\ \hline
\textbf{Evaluation Metrics}            & $F_{\text{micro}}$ & $F_{\text{macro}}$ & $F_{\text{micro}}$ & $F_{\text{macro}}$ & Score  \\ \hline
\textbf{nevermore}             & \textbf{0.958} & \textbf{0.781} & \textbf{0.962} & \textbf{0.836}    & 77.57  \\ \hline
\textbf{jiachx}             & 0.952 & 0.748 & 0.958 & 0.815    & 69.64  \\ \hline 
\textbf{xlzhang}              & 0.952 & 0.760 & 0.958 & 0.811 & 69.64 \\ \hline 
\textbf{HFL}                 & 0.953 & 0.769 & 0.958 & 0.811    & \textbf{77.70}  \\ \hline
\textbf{大师兄}             & 0.945 & 0.757 & 0.951 & 0.816    & 73.16   \\ \hline
\textbf{安徽高院类案指引研发团队} & 0.946 & 0.756 & 0.950 & 0.803   & 72.24\\ \hline
\textbf{AI\_judge}          & 0.952 & 0.766 & 0.956 & 0.811    & -- \\ \hline
\textbf{只看看不说话}       & 0.948 & 0.738 & 0.954 & 0.801    & 77.54 \\ \hline
\textbf{DG}                    & 0.945 & 0.717 & 0.949 & 0.755    & 76.18 \\ \hline
\textbf{SXU\_AILAW}         & 0.940 & 0.728 & 0.950 & 0.791    & 76.49 \\ \hline
\textbf{中电28所联合部落}   & 0.934 & 0.740 & 0.937 & 0.772    & 75.77 \\ \hline
\end{tabular}
\caption{Performance of participants on \contest.}
\label{table:perf_rc_sys}
\end{table*} 

There are over $200$ teams who have registered for CAIL2018 and submitted their final models. The final scores show that neural models can achieve considerable results on task $1$ and task $2$, but it is still challenging to predict the prison terms. In Table~\ref{table:perf_rc_sys}, we list the scores of top-$6$ participants of each subtask. Here, we evaluate these models on the testing set in the second stage, which contains $35,922$ cases. 

We have collected the technical reports of these contestants. In the following parts, we summarize their methods and tricks according to these reports.

\subsection{General Architecture}
\textbf{Pre-processing.} For most contestants, they conduct the following pre-processing steps to transform the raw documents into the format which is suitable for their models.

\begin{itemize}
    \item \emph{Word Segmentation}. As all the documents are written in Chinese, it is important for the contestants to conduct a high-quality word segmentation. For word segmentation, the contestants usually choose jieba\footnote{\url{https://github.com/fxsjy/jieba}}, ICTCLAS\footnote{\url{http://ictclas.nlpir.org/}}, THULAC\footnote{\url{http://thulac.thunlp.org/}} or other Chinese word segmentation tools.
    

    \item \emph{Word Embedding}. After word segmentation, we need to transform the discrete word symbols into continuous word embeddings. Generally, the contestants employ word2vec~\cite{mikolov2013distributed}, Glove~\cite{pennington2014glove}, or FastText~\cite{joulin2017bag} to pre-train word embeddings on these criminal cases.
    \end{itemize}

\textbf{Text Classification Models}. After preprocessing, we need to classify these processed fact descriptions into corresponding categories. For most contestants, they employ existing neural network based text classification models to extract efficient text features. The most commonly used text classification models are listed as follows:

\begin{itemize}
    \item \emph{Text-CNN}~\cite{kim2014textcnn}: CNN with multiple filter widths.
    \item \emph{LSTM}~\cite{hochreiter1997long}) or bidirectional LSTM.
    \item \emph{GRU}, Gated Recurrent Unit~\cite{Cho2014Learning}.
    \item \emph{HAN}, Hierarchical Attention Networks~\cite{yang2016hierarchical}.
    \item \emph{RCNN}, Recurrent Convolutional Neural Networks~\cite{lai2015recurrent}.
    \item \emph{DPCNN}, Deep Pyramid Convolutional Neural Networks~\cite{johnson2017deep}.
\end{itemize}

According to the technical reports of contestants, it has been proven that these neural models can achieve good performance in high-frequency categories.





\subsection{Promising Tricks}
In predicting relevant law articles and charges, these traditional models can achieve promising results in high-frequency categories. However, due to the imbalance issue, it is challenging to reach a good performance on the low-frequency ones. Therefore, how to address the problem of imbalanced data becomes the most important thing in the first two subtasks.

In the task of predicting prison terms, simple linear regression methods perform poorly than classification models. Thus, most participants still treat it as a text classification problem. However, how to divide the intervals is challenging and will badly influence the classification performance. Meanwhile, the prison terms are affected by many factors and explicit features, rather than implicit semantic meanings in the text. All these issues make the task $3$ the most difficult subtask.

According to the technical reports, there are some useful tricks which can address these issues and improve the text classification models significantly. We summarize them as follows:


\begin{itemize}
    \item \textbf{Word Embeddings.} It has been proven by participants that a better word embedding model, such as ELMO~\cite{peters2018deep} could achieve a better performance than Skip-Gram\cite{mikolov2013distributed}. Moreover, training word embeddings on a larger legal corpus can also improve the performance of LJP models.
    \item \textbf{Data Balance.} Undersampling and oversampling methods are the most common ways to address the imbalance issue of categories in this competition.
    \item \textbf{Joint Learning.} As there are dependencies among these subtasks, some participants employ multi-task learning models to solve them jointly.
    \item \textbf{Additional Attributes.} Inspired by \citet{niubi2018coling}, participants improve their performance on few-shot and confusing category pairs by predicting their legal attributes.
    
    \item \textbf{Additional Features.} Many participants attempted to extract features manually, e.g., amount involved, named entities, ages and so on. These manually defined features can improve the performance of task $3$ greatly.

   \item \textbf{Loss Function.} Most models use cross-entropy as their loss functions. However, some models adopt more promising loss functions, such as focal loss~\cite{lin2018focal}, to enhance the performance on low-frequency categories. Besides, the loss weights of various categories and the activation functions of the output layer also have great influence on the final performance.
   \item \textbf{Ensemble.}  Most participants train several different classification models and combine them with simple voting or weighted average strategies to combine their predicting results.
    
\end{itemize}

\subsection{Conclusion}

In CAIL2018, we employ Legal Judgement Prediction as the competition topic. In this competition, we construct and release a large-scale LJP dataset. The performance of $3$ LJP subtasks significantly raised with the efforts of over $200$ participants. In this paper, we summarize the general architecture and promising tricks they employed, which are expected to benefit further researches on legal intelligence.


\bibliographystyle{acl_natbib_nourl}
\bibliography{reference}

\end{CJK*}

\end{document}